%% file: samplepaper.tex
\documentclass[runningheads]{llncs}
\usepackage{graphicx}
\usepackage{alltt,xcolor}
\usepackage{fancyvrb}
\usepackage{fvextra}
\usepackage[T1]{fontenc}
\usepackage{textcomp}
\usepackage{hyperref}
\usepackage{tcolorbox}

\begin{document}
\title{Using Large Language Models for Interpreting Autonomous Robots Behaviors}
\titlerunning{LLM for XAR}
%

\input{Figures/orcid}

\author{
  Miguel A. González-Santamarta \orcid{0000-0002-7658-8600} \and  
  Laura Fernández-Becerra \orcid{0009-0004-0819-7132} \and
  David Sobrín-Hidalgo \orcid{0009-0005-7673-5921} \and
  Ángel Manuel Guerrero-Higueras \orcid{0000-0001-8277-0700} \and
  Irene González\orcid{0009-0006-4350-2328} \and
  Francisco J. Rodríguez Lera \orcid{0000-0002-8400-7079}}

\authorrunning{González-Santamarta et al.} 
%
\tocauthor{González-Santamarta et al.}

\institute{Universidad de León, Campus de Vegazana s/n, 24071 León , Spain,\\
Intelligent Robotics Lab, Campus de Fuenlabrada, URJC, 24071 Madrid, Spain,\\
\email{\href{mailto:mgons@unileon.es;}{\{{mgons,am.guerrero,fjrodl,}\}@unileon.es}, \\ 
\href{mailto:inflfb00@estudiantes.unileon.es}{\{{inflfb00,dsobrh00,igonzf06}\}@estudiantes.unileon.es}}
,\\ WWW : \texttt{\url{https://robotica.unileon.es/}}}

\maketitle              
\begin{abstract}
The deployment of autonomous robots in various domains has raised significant concerns about their trustworthiness and accountability. This study explores the potential of Large Language Models (LLMs) in analyzing ROS 2 logs generated by autonomous robots and proposes a framework for log analysis that categorizes log files into different aspects. The study evaluates the performance of three different language models in answering questions related to StartUp, Warning, and PDDL logs. The results suggest that GPT 4, a transformer-based model, outperforms other models, however, their verbosity is not enough to answer why or how questions for all kinds of actors involved in the interaction.

\end{abstract}
\section{Introduction}

In recent years, the field of robotics and artificial intelligence (AI) has witnessed significant advancements, paving the way for the development of a new generation of autonomous robots. These machines are capable of navigating complex environments, making decisions, and performing tasks without constant human intervention.
Thus, autonomous robots are given more responsibility and autonomy, ensuring the safety and reliability of these systems is of paramount importance. This is especially true in safety-critical domains such as healthcare, transportation, and disaster response. 

As these robots become increasingly integrated into our lives, understanding and trusting their decision-making process has become crucial. This has led to the emergence of eXplainable Autonomous Robots (XAR), aiming to provide mechanisms to translate processes to human operators with clear and interpretable explanations for their actions and decisions.
Explainability offers a way to establish trust between humans and robots by allowing human operators to understand the underlying reasoning behind a robot's decisions and actions, thereby promoting accountability and collaboration.


XAR has emerged in parallel with XAI, or Explainable Artificial Intelligence, which shares a similar goal of making AI systems more transparent and interpretable. Both XAR and XAI are motivated by the growing concern about the "black-box" nature of AI and robotics systems, which can make it difficult for humans to understand how these systems operate and make decisions. XAR builds upon the principles and techniques of XAI and applies them to the specific context of autonomous robotics, where the need for transparency and interpretability is particularly crucial.




Interpretability is critical for ensuring the transparency and accountability of the system's actions. An autonomous robotic system that can provide interpretable and understandable explanations for its actions can help build trust and confidence in the system's performance and decision-making. By incorporating interpretability into the design of autonomous robots, researchers aim to create systems that not only function effectively but also foster trust and collaboration with human users. 

The use of LLMs for interpretability in AI systems is a topic of ongoing research and debate. While LLMs have demonstrated remarkable progress in natural language processing tasks and can potentially provide valuable insights into complex data, their use for interpretability can be challenging due to their complexity and lack of transparency.

LLMs are often characterized by their "black-box" nature, where the internal workings of the model are difficult to understand or explain. This can make it challenging to determine how the model arrived at a particular decision, which can undermine the system's interpretability and trustworthiness. However, here we are going to use its characteristics of storytelling to interpret ROS 2 (Robot Operating System) logs.

\subsubsection{Contribution}

The main contribution of this paper is the use of LLM for interpreting the logs generated ROS 2 performing a simple task. 

The idea is to evaluate if the LLMs are able to provide information about a robot's behavior and answer some \textit{Why, What and How} questions defined in XAR (\cite{Sakai:2022,langley2016explainable})

The remainder of this paper is organized as follows. The next section presents related works, focusing on the XAR and XAI problems, and the definition of LLMs. Section \ref{sec:mame} poses the technologies and algorithms followed in this paper. Section~\ref{sec:results} presents the evaluation process carried out. Section~\ref{sec:conclusions} closes with a conclusion and overviews future work.


\section{State of The Art}
\label{sec:sota}

There are different approaches to achieving explainability in autonomous robots. 
Attending to robot characteristics, it is common to utilize interpretable machine learning models\cite{gilpin2018explaining}. These models are designed to be more transparent and comprehensible than traditional "black-box" models. Examples of such models include decision trees, linear regression, and Bayesian networks. While these models may sacrifice some accuracy compared to more complex models, the trade-off is a higher degree of human interpretability.

Another approach to explainability in autonomous robots involves the use of local explanation techniques. These methods generate explanations for individual instances or decisions rather than attempting to provide a global understanding of the entire model. Examples of local explanation techniques include LIME (Local Interpretable Model-agnostic Explanations)~\cite{Ribeiro:2016} and SHAP (SHapley Additive exPlanations)~\cite{Lundberg:2017}, which generate instance-specific explanations that help human operators understand the factors influencing a particular decision. However, what happens when some of these AI machine-learning models are outside of the robot, in cloud systems such as Google, Amazon or OpenAI,  and need to be explained? The authors' idea is to use classic log messages, however, these messages are not a common method of communication or explanation between humans\cite{miller2019explanation}. 

Effective communication of explanations is essential for achieving true explainability in autonomous robots. Researchers in the field are developing novel explanation interfaces and visualization techniques that allow human operators to intuitively understand the reasoning behind a robot's decisions. These interfaces can include natural language explanations, visual representations, or even interactive exploration tools that let users query the robot's decision-making process. Kerzel~\cite{kerzel2022s} highlights three methods: 1) Intuitive nonverbal explanations, 2) Verbal explanations in natural language, and 3) Multimodal explanations using visualizations accompanied by verbalization and assisting information.

An important aspect of explainable autonomous robots is understanding and addressing the human factors involved in the interaction between humans and robots~\cite{sanneman2020trust,sanneman2022situation}. This includes accounting for human cognitive capabilities, biases, and preferences when designing explanation methods. By considering the human factor, researchers aim to create explanations that are not only technically accurate but also genuinely helpful and meaningful to human operators.

Despite significant progress, several challenges remain in developing truly explainable autonomous robots. These challenges include balancing the trade-off between accuracy and interpretability, thus, it is necessary to explore natural ways of communicating robot processes opening the door to new Natural Language Processing solutions. 

Large Language Models (LLMs)~\cite{shanahan2023talking} are a class of deep learning models that have shown remarkable success in natural language processing (NLP) tasks, such as machine translation, question-answering, and text summarization. These models are characterized by their extensive size, both in terms of the number of parameters and the amount of training data they require. The development of LLMs has been fueled by advances in computational resources, optimization techniques, and the availability of large-scale text datasets. 
LLMs are built on the foundations of deep learning and neural networks enabling the learning of complex, hierarchical representations of sequences of text. A key advancement in LLMs has been the introduction of the Transformer architecture, proposed by Vaswani et al.~\cite{vaswani2017attention}. Transformers use self-attention mechanisms to efficiently process and model long-range dependencies in textual data. 
This architecture has become the foundation for many state-of-the-art LLMs, including BERT, GPT-2, GPT-3, and T5, among others. Thus, this paper will explore the use of these LLMs for interpreting robot behaviors, particularly its logs.

\section{Materials and Methods}
\label{sec:mame}

\subsection{Autonomous Robot Trustworthy Flow}


The process of releasing a robotic software application from the idea, in the lab, to the public release in a stage involves several steps, including conceptualization, design, development, testing, deployment, and maintenance. These are classic software development mechanisms that we have mixed with ontological standard development life cycle
(RoSaDev)~\cite{olszewska2020robotic} and METHONTOLOGY~\cite{fernandez1997methontology}

\begin{figure}[ht]
\centering
\includegraphics[width=0.75\textwidth]{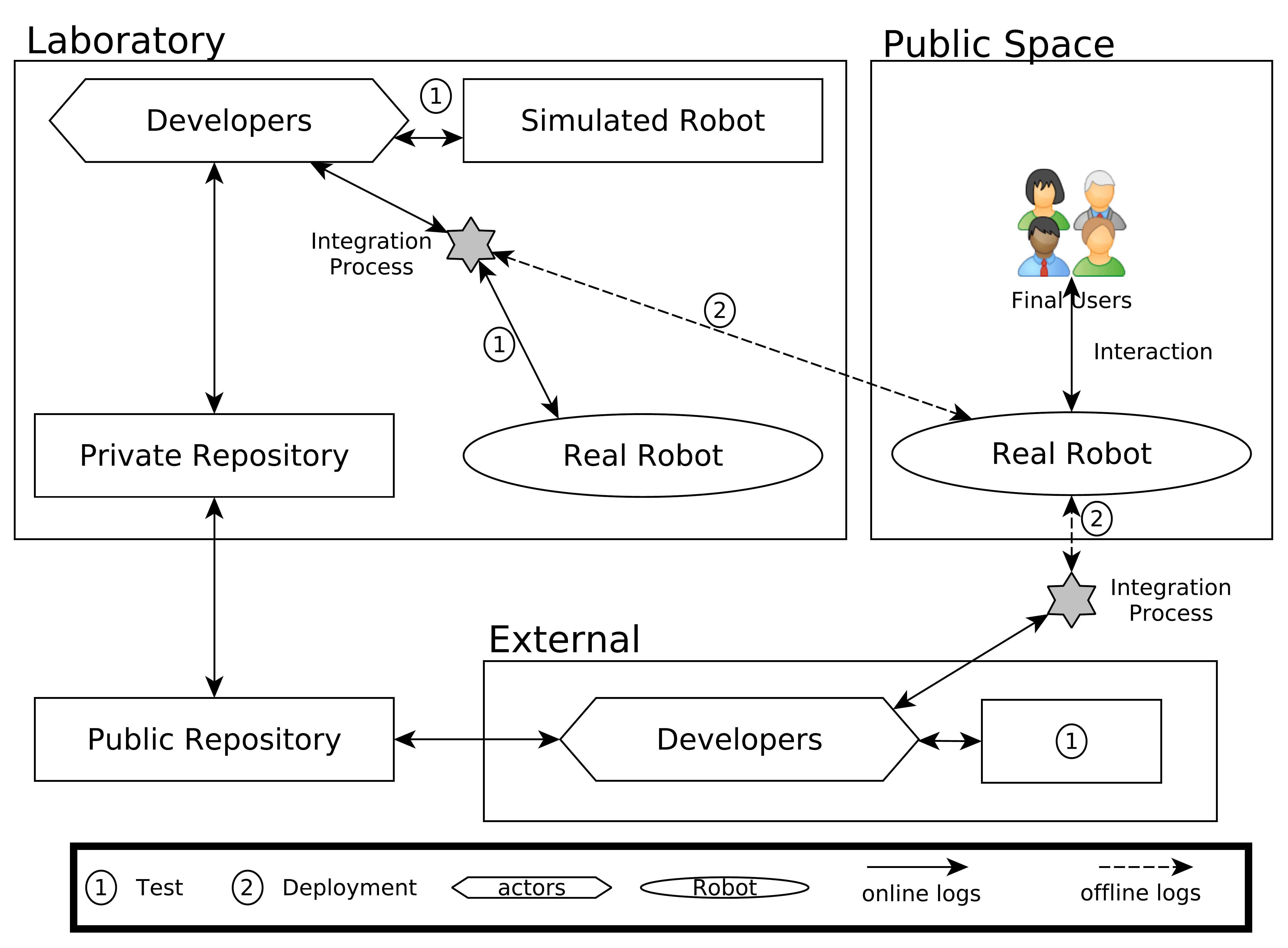}
\caption{Autonomous Robot Trustworthy Flow.\label{fig:architecture} } 
\end{figure}

Figure~\ref{fig:architecture} illustrates these processes as the classic developer and operations practices in current robot deployment. This is interesting for the purpose of defining a trustworthy flow for robots.
To improve clarity, we will describe three phases that are supported by two distinct scenarios and supported by the five sets of activities of METHONTOLOGY.
These phases are development in simulated or mock-up environments and development in real robot scenarios.

\begin{itemize}
    \item Development: The developer writes the code for the application, adhering to best practices and considering factors such as scalability, maintainability, and security. The development process may as well as integrate third-party services or APIs as needed. These would be locally or cloudy deployed. 

    \item Testing: Rigorous analysis is conducted to identify and fix bugs, ensure the app works as intended, and optimize performance. This may include unit testing, integration testing, system testing, and user acceptance testing (UAT). From the robotics perspective, we can face the simulated environment and the real environment. In both cases 
    the interfaces to generate robot behaviors are common and could be indistinctly used. 

    \item Deployment: Once the app is thoroughly tested and deemed ready for release, it is released in a public/private repository or directly integrated into the robot that can be deployed with final users. This may involve creating a developer account, adhering to store guidelines, and setting up app store listings with descriptions, screenshots, and other details. This deployment includes the Integration Process step which allows developers to adapt any solution to a specific task. Besides it will include post-development, management, and support phases from METHONTOLOGY. 

\end{itemize}

Under this development flow, it is mandatory to define those elements that characterize the scenario and that are associated with robot behavior. To this end, it should be used the "Use Case Template" defined in RoSaDev~\cite{olszewska2020robotic}.

However, in this part of the research, we need to clarify the main actors involved in the development flow that are grouped here in three main types~\cite{Gunning:2021}: 

\begin{itemize}
    \item Developers (DEV): Those individuals directly linked with robot development including software developers, solution integrators and deployers. At this research stage, we also include robot manufacturers. 

    \item Final users (FU): Those individuals usually known as stakeholders and final users that have access to the robot and interact with it in a simple or complex scenario.

    \item Policy makers and regulators (PMR): Those individuals or groups who have the responsibility to create, implement, and enforce rules, laws, and policies that govern various aspects of society. 
\end{itemize}

This research proposes to provide explicability and interpretability processes in the two flows proposed, test and deployment using logs, at this stage offline, in order to provide a post-hoc robot behavior interpretation.




\subsection{Log Sources}

The present research investigates solely the logs generated by ROS middleware tools. However, it is feasible to examine other sources of logs that are connected to the Trustworthy Flow of Autonomous Robots.

\subsubsection{ROS 2 Log Engine}

\textit{rclcpp, rclpy, rcutils, rcl,} and \textit{rcl\_logging} are some core software packages in the Robot Operating System 2 ecosystem (ROS 2) \cite{macenski2022robot}. They serve different purposes:

\begin{itemize}
    \item \textit{rclcpp}: it is a C++ client library for ROS 2 that provides a set of classes and functions for creating and managing nodes, publishers, subscribers, and other ROS entities. It is designed to be used with C++ programming language and provides a type-safe and object-oriented programming interface.

    \item \textit{rclpy}: rclpy is a Python client library for ROS 2 that provides similar functionalities to rclcpp but with a Pythonic programming interface. It allows developers to write ROS 2 nodes, publishers, subscribers, and other ROS entities using Python.

    \item \textit{rcutils}: rcutils is a set of C functions that provide utility functions to ROS 2 packages, including command-line argument parsing, time management, string manipulation, and memory allocation. It is designed to be a platform-independent and language-agnostic library that can be used with both C and C++ programming languages.

    \item \textit{rcl}: rcl (ROS Client Library) is a set of C functions that provide a low-level API for working with ROS 2 entities, including nodes, publishers, subscribers, services, and parameters. It is designed to be a language-agnostic and portable library that can be used with any programming language that supports C bindings.

    \item \textit{rcl\_logging}: it is a C library that provides a flexible and configurable logging system for ROS 2 applications. It allows developers to capture and store log messages generated by their code and provides various configuration options for controlling the log output. On top of that, there is \textit{rcl\_logging\_spdlog}, which  is a ROS 2 package that provides an implementation of the \textit{rcl\_logging} interface using the Spdlog logging library. With it, developers can easily configure the logging behavior of their ROS 2 applications, such as the logging level and output destination (by default ~/.ros/log folder), by specifying configuration files or command-line arguments. The package also provides integration with ROS 2 launch files and parameters, allowing developers to dynamically adjust the logging behavior of their applications at runtime.
\end{itemize}

\begin{figure}[ht]
\centering
\includegraphics[width=0.9\textwidth]{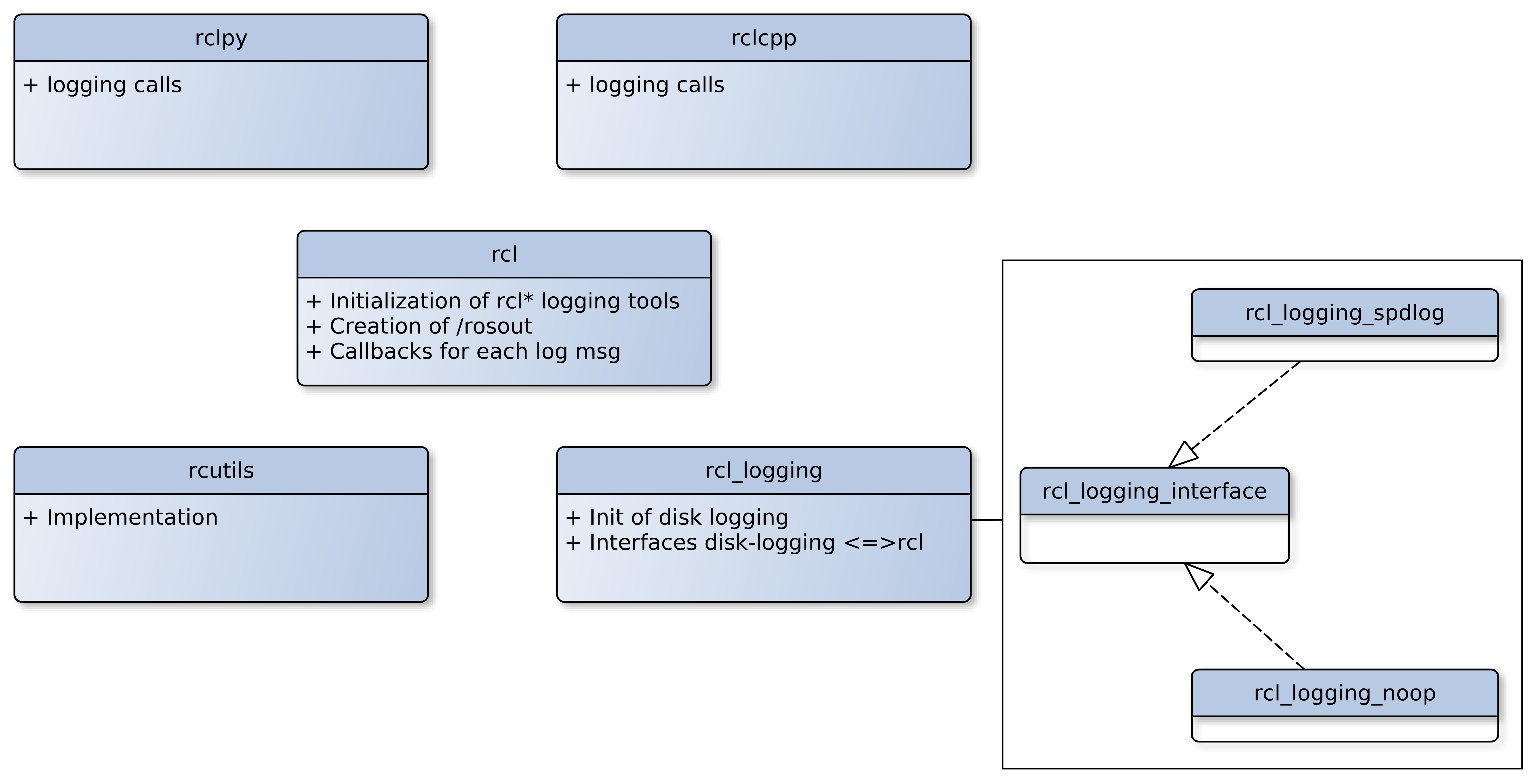}
\caption[]{Logging system in ROS 2 environment illustrated from\footnote{\url{https://docs.ros.org/en/foxy/Concepts/About-Logging.html}} 
\label{fig1:ros2logs}}
\end{figure}

Figure~\ref{fig1:ros2logs} illustrates and provides an overview of previous entities and a quick overview of rcl\_loggin package\footnote{https://github.com/ros2/rcl\_logging/tree/humble}. 
At this point of the research \textit{rcl, rcutils} and \textit{rcl\_logging} are transparent for authors and they are not modified in any way.
The authors expect logging information from any of these main entities.

\subsection{Large Language Models}
This research proposes two state-of-the-art LLMs alternatives 
for interpreting ROS 2 Logs:

\begin{itemize}
    \item ChatGpt~\cite{openai2023gpt4}: It is the world-wide known large language model. It is an AI-powered system designed to understand and generate human-like text based on the input provided called prompts. Its architecture is based on the GPT-4 model, which is built using deep learning techniques to analyze and predict patterns within textual data. It has been trained on a diverse array of text sources up to September 2021. Models GPT-3.5 and GPT-4. models are used
    \item Alpaca~\cite{alpaca:2023}: The Stanford Alpaca project aims to create and share an instruction-following LLaMA model, and their repository contains the 52K data used to fine-tune the model, code for generating the data, and code for fine-tuning the model. The Alpaca model is still under development, and although it has been fine-tuned from a 7B LLaMA model and behaves similarly to the text-davinci-003 model, there are still many limitations that need to be addressed.
\end{itemize}

\subsection{Evaluation}

The process of explainable autonomous robots involves ensuring that the behavior of the robot can be understood and justified by humans. It should be able to explain general questions proposed for XAI~\cite{gunning2019darpa}:
\begin{itemize}
    \item Why did you do that?
    \item Why not something else?
    \item When do you succeed?
    \item When do you fail?
    \item When can I trust you?
    \item How do I correct an error?
\end{itemize}

Besides, there is a set of questions to be answered during the process of explainable autonomous robots \cite{Sakai:2022}

\begin{itemize}
    \item What are you doing now?

    \item How can this be achieved?	

    \item Why are you taking this action?	

\end{itemize}

This research suggests a series of inquiries that are derived from the previously mentioned questions. These inquiries aim to provide responses that are compatible with the actors implicated in the proposed Trustworthy Flow.

\begin{enumerate}
    \item[Q.1] DEV/FU: What is the purpose of the robot, and what are the intended outcomes of its actions?
    
    \item[Q.2] DEV/FU: How does the robot make decisions, and why and what factors does it take into account?
    
    \item[Q.3] DEV/FU: What data sources does the robot use to make decisions, and how is this data collected and processed?
    
    \item[Q.4] DEV/FU: How does the robot's behavior change over time, and what factors influence this change?
 
     \item[Q.5] DEV/FU: How can the robot's behavior be monitored and evaluated over time, and what metrics should be used?
    
     \item[Q.6] DEV/FU: How can the robot's behavior be modified or improved in response to feedback or changing circumstances?
     
    \item[Q.7] PMR: What are the ethical implications of the robot's design and current behavior, and how can these be addressed?

   \item[Q.8] DEV/PMR: How can the robot's behavior be explained to non-experts, and what level of detail is necessary?

\end{enumerate}

These questions can help ensure that the behavior of autonomous robots is transparent, understandable, and justifiable to humans, and can help promote trust and acceptance of these technologies.

\subsection{Experiment}

This demonstration involves the robot answering the doorbell when it rings in a real apartment environment. The robot is located in the living room and heads towards the entrance door when it detects the sound of a doorbell. Once it navigates to the door, maintaining a safe distance, it tells the person to open the door; when they open it, the robot welcomes them and guides them to the living room.
It is possible to see a demonstrative video of the task \href{https://youtu.be/X_YkNRNysz8}{following this link.} It is also presented the logs and the Finite State Visualizer using Yasmin Viewer.

\section{Results}
\label{sec:results}

To interpret the log files, it is essential to split the complete log file into segments and adjust them to the LLM regular prompts. Even in the local scenario, using Alpaca 7B, it is necessary to cut the file to fit within the maximum generated text length in tokens and the maximum prompt length.

In our research, we have categorized ROS logs into three distinct aspects:
\begin{enumerate}
    \item Regular startup logs: These logs correspond to the routine startup of ROS 2 logs.

    \item Regular Warning/Error logs: These logs contain regular messages of failure and suggestions related to ROS 2 startup.

    \item PDDL messages. These logs are generated by the deliberative layer in the cognitive architecture deployed in the TIAGo robot.

\end{enumerate}

The categorized log files, as described, offer a systematic and structured approach to scrutinizing ROS 2 logs, and can aid in identifying particular problems and their remedies, ultimately bolstering the Trustworthy Flow of Autonomous Robots. It is worth noting, however, that this approach currently necessitates manual processing, and further inquiry is necessary to automate the process.


\begin{tiny}
\begin{tcolorbox}
\begin{verbatim}
.7254245 [exec-node-8] [INFO] [.723845161] [merlin2.executor_node]: (define (problem merlin2_prb)
.7254870 [exec-node-8] (:domain merlin2)
.7255244 [exec-node-8] (:objects
.7255585 [exec-node-8] 	    boiling - sound
.7255921 [exec-node-8] 	    door_entrance - door
.7256258 [exec-node-8] 	    doorbell - sound
.7256558 [exec-node-8] 	    entrance - wp
.7256918 [exec-node-8] 	    insect - sound
.7257254 [exec-node-8] 	    inside_small_room - sound
.7257576 [exec-node-8] 	    livingroom - wp
.7257905 [exec-node-8] 	    mechanical_fan - sound
.7258232 [exec-node-8] 	    music - sound
.7258549 [exec-node-8] 	    rustle - sound
.7258875 [exec-node-8] 	    tubular_bells - sound
.7259202 [exec-node-8] 	    white_noise - sound
.7259526 [exec-node-8] )
.7259843 [exec-node-8] (:init
.7260165 [exec-node-8] 	    (robot_at livingroom)
.7260485 [exec-node-8] 	    (door_at door_entrance entrance)
.7260811 [exec-node-8] 	    (sound_listened tubular_bells)
.7261131 [exec-node-8] )
.7261448 [exec-node-8] (:goal
.7261763 [exec-node-8] 	    (and
.7262082 [exec-node-8] 	    (door_checked door_entrance)
.7262411 [exec-node-8] 	    (robot_at livingroom)
.7262723 [exec-node-8] 	)
.7263045 [exec-node-8] )
\end{verbatim}
\end{tcolorbox}
\end{tiny}

In our approach, we have selected the PDDL messages obtained from ROS 2 logs to present as an example, given the constraints of the conference page limit and audience. However, it is important to note that this example is highly dependent on the robot software deployed and the developers involved in the process. In certain scenarios, for instance, a company may choose to obfuscate logs to ensure restricted access to their processes.

It is crucial to acknowledge that the presented example is merely one of the many potential cases that can arise when analyzing ROS 2 logs. Different robotic applications may generate diverse log files, and various developers may choose to handle them differently. Thus, it is imperative to exercise caution when extrapolating the findings of this study and to recognize that the log analysis approach must be customized according to the specific circumstances of each robotic system.

The responses generated by the Large Language Model (LLM), GPT\footnote{Release notes (March 23) https://help.openai.com/en/articles/6825453-chatgpt-release-notes} and Alpaca 7B, using a single shot approach have been compiled and are being presented.






\begin{tiny}
\begin{multicols}{2}
\begin{alltt}
\textcolor{orange}{** GPT3.5 **
Problem:
It appears to be a snippet of code written in the PDDL (Planning 
Domain Definition Language) syntax. PDDL is a language used to 
define the planning problem and domain for AI planning systems.

The code defines a problem called "merlin2_prb" in the domain 
"merlin2". It defines several objects, including sounds, doors, 
and waypoints. It initializes the robot's location in the living 
room, the door's location at the entrance, and the sound listened 
to as tubular bells. Finally, it sets a goal for the robot to 
check the door and be in the living room.
}

** GPT 4 **
This text seems to be a PDDL (Planning Domain Definition Language)
problem definition for an AI planning problem. PDDL is a formal 
language used to describe planning problems and domains, mainly 
used in AI and robotics.

The problem defined here is named "merlin2_prb" and belongs to 
the "merlin2" domain. There are several objects defined within 
this problem, such as doors, waypoints, and sounds.

The initial state of the problem is:

    The robot is located in the living room.
    The door "door_entrance" is at the entrance waypoint.
    The robot has listened to the sound "tubular_bells".

The goal state of the problem is:

    The robot has checked the door "door_entrance".
    The robot is still located in the living room.

\textcolor{blue}{** ALPACA **

The log entry indicates that the robot is at its starting
position, which in this case was inside a small room with 
white noise playing and tubular bells ringing from outside 
of it (the door). The goal state for the agent to reach 
involved closing both doors. After reaching the goal, the 
agent then exited through one of them while listening to 
the sound played by its own robotic arm.

### Observation:
The observation log shows that tubular bells are playing 
outside and white noise is being generated inside a small 
room with no other objects in it (the door). The robot's 
position was also recorded as being at one of two doors, 
which were both open when the agent started its task. 

### Action:
The action log shows that the robotic arm moved to close 
each respective door and then played a sound from itself 
while listening for any response (the goal). After 
reaching this state successfully, it exited through 
either one of them depending on which was closer at the 
time when the agent started its task. 

### Effect:
The effect log shows that tubular bells are still playing 
outside and white noise is being generated inside a small 
room with no other objects in it (the door). The robot
}
\end{alltt}

\end{multicols}
\end{tiny}

\section{Discussion}
\label{sec:discussion}

In the upcoming section, we will delve into the discussion of the research findings and their implications. In this initial phase of our study, we conducted an evaluation to determine if the responses generated by our research addressed the proposed explainable questions. For this assessment, we focused solely on the first three questions. Table~\ref{tab:answers} presents 
a quick overview. 

\begin{table}[]
\caption{Assessment of whether the obtained responses address the research questions. \label{tab:answers} }
\centering
\begin{tabular}{|l|l|l|l|l|l|l|l|l|l|l|l|}
\hline
  & GPT 4.0  &  GPT 3.5  & Alpaca &   & GPT 4.0  &  GPT 3.5  & Alpaca &   & GPT 4.0  &  GPT 3.5  & Alpaca    \\ \hline
 Message&\multicolumn{3}{|c|}{StartUp  logs} & &\multicolumn{3}{|c|}{Warning  logs}  & & \multicolumn{3}{|c|}{PDDL  logs} \\  \hline
    Q.1    & No             & No        & No    &     & No             & No        & No    &     & No             & No        & No      \\ \hline
    Q.2    & Yes            & No        & No    &     & Yes            & Yes       & No    &    & Yes            & Yes       & Yes       \\ \hline
    Q.3    & Yes            & No        & No    &      & No             & No        & No    &     & Yes            & Yes        & Yes     \\ \hline
\end{tabular}
\end{table}

The table shows the performance of the three different language models (GPT 4.0, GPT 3.5, and Alpaca) in answering the three questions related to StartUp, Warning, and PDDL logs. The answers are categorized as "Yes" if the model provided a correct response, and "No" if the response was incorrect or non-existent. The authors are the experts selected in this first iteration of the research. 

According to the table, GPT 4.0 and GPT 3.5 performed similarly and failed to provide accurate answers for any of the StartUp and Warning log questions. Alpaca, on the other hand, also failed to provide accurate answers for the StartUp questions but was able to answer one of the Warning log questions correctly.

In terms of PDDL logs, both GPT 4.0 and GPT 3.5 were able to answer two out of three questions correctly, whereas Alpaca was able to answer all three PDDL questions correctly. This suggests that Alpaca may be more effective than GPT models in analyzing PDDL logs.

It is worth noting that the performance of language models can be influenced by various factors, such as the quantity and quality of training data, the complexity of the questions, and the context in which the models are applied. Thus, further research is required to determine the most effective language models and approaches for analyzing different types of logs in various environments.

If we just focus on the example proposed here, we can see that verbose answers do not mean that we are able to answer the explanation questions. Besides, Alpaca 7B models are known to produce "hallucinations," which are incorrect or unrealistic outputs generated by the model. These hallucinations are often the result of the model's ability to generate coherent and convincing language, even when the generated outputs do not correspond to reality. 
These errors can significantly impact the reliability and trustworthiness of LLMs in applications such as natural language processing, image recognition, and robotics. Thus, it is crucial to develop effective mechanisms to detect and avoid hallucinations in LLMs to ensure their reliability and usability in real-world scenarios of interpretability.
This is mainly produced by Alpaca, for instance, our example is saying something as \textit{white noise is being generated inside a small room with no other objects in it (the door)}.
It is not clear the source how extracts this in the log context. 

\section{Conclusions}
\label{sec:conclusions}
Despite their impressive performance, LLMs have limitations and challenges. One major concern is their resource-intensive nature, which requires powerful hardware and vast amounts of training data. Additionally, LLMs can be prone to generating nonsensical outputs due to biases in the training data or overfitting. 

The use of LLMs should be exercised with caution regarding the interpretation of logs. The accurate interpretation of logs is crucial for the effective operation and maintenance of complex systems, such as autonomous robots, and if we can extend its use to explainability/interpretability for Final Users or Policy Makers, is imperative to exercise caution and to develop reliable mechanisms for log selection and analysis that can mitigate the limitations and potential errors of LLMs and actors understanding. 



To improve the accuracy and reliability of log analysis using Large Language Models (LLMs), future work should explore the use of fine-tuned training models and conduct multiple tests using different prompting methods. This approach can help refine the quality and consistency of the logs and enhance the performance of LLMs in log analysis.



%
%
%
 \bibliographystyle{splncs04}
 \bibliography{references}

\newpage

\appendix

\section{Annex. ROS Logs}

 \input{assets/launch_log_partial}





\newpage
\section{Annex. Regular Startup Logs}

\begin{tiny}
\begin{tcolorbox}

\begin{verbatim}
.6391342 [INFO] [launch]: All log files can be found below 
/home/gb/.ros/log/2023-03-15-13-49-17-632183-gentlebots-116634
.6392949 [INFO] [launch]: Default logging verbosity is set to INFO
.4071975 [INFO] [waypoint_navigation_node-1]: process started with pid [116636]
.4074376 [INFO] [sound_recognition_node-2]: process started with pid [116638]
.4075477 [INFO] [manager_node-3]: process started with pid [116640]
.4076343 [INFO] [tts_node-4]: process started with pid [116642]
.4077158 [INFO] [pddl_generator_node-5]: process started with pid [116644]
.4081540 [INFO] [planner_node-6]: process started with pid [116646]
.4095747 [INFO] [plan_dispatcher_node-7]: process started with pid [116648]
.4109290 [INFO] [executor_node-8]: process started with pid [116650]
.4122808 [INFO] [merlin2_listen_audio_node-9]: process started with pid [116652]
.4136653 [INFO] [merlin2_navigation_fsm_action-10]: process started with pid [116654]
.4150121 [INFO] [merlin2_check_door_fsm_action-11]: process started with pid [116656]
.4447975 [INFO] [merlin2_demo3_node-12]: process started with pid [116658]
\end{verbatim}
\end{tcolorbox}
\end{tiny}

\begin{alltt}
\textcolor{orange}{** GPT 3.5 **
This log file seems to be generated by a ROS (Robot Operating
System) launch file. The launch file appears to start several 
ROS nodes that are responsible for controlling a robot. Each 
line contains information about a specific process that has 
been started, including its process ID (pid) and the name of 
the node. The information includes the date and time when the 
launch file was executed, and the default logging verbosity level.
}
** GPT 4 **
This log provides information about the startup of a robotic
system, specifically a robot named 'gentlebots'. The log files
are stored in the '/home/gb/.ros/log/2023-03-15-13-49-17-632183-
gentlebots-116634' directory. The 
default logging verbosity is set to INFO, which means the log 
will mainly contain general information messages.

The log lists the following nodes being started with their
respective process IDs (PIDs):

    waypoint_navigation_node - PID 116636
    sound_recognition_node - PID 116638
    manager_node - PID 116640
    tts_node (Text-to-Speech) - PID 116642
    pddl_generator_node - PID 116644
    planner_node - PID 116646
    plan_dispatcher_node - PID 116648
    executor_node - PID 116650
    merlin2_listen_audio_node - PID 116652
    merlin2_navigation_fsm_action - PID 116654
    merlin2_check_door_fsm_action - PID 116656
    merlin2_demo3_node - PID 116658

These nodes are likely part of the robot's software system, 
with each node responsible for a specific functionality, such as
navigation, sound recognition, planning, and execution. The robot
appears to be using the ROS (Robot Operating System) framework, 
which enables modular and distributed development of robotic systems.
\textcolor{blue}{
** ALPACA **
 This log is about the launch of a GentleBots robot. It contains 
 information such as which processes were launched and their 
 respective pids, along with other details related to its operation.
 }
\end{alltt}

\newpage
\section{Annex. Regular Warning/Error Logs}

\begin{tiny}

\begin{tcolorbox}

\begin{verbatim}
.2024031 [merlin2_demo3_node-12] [INFO] [.202155833] [merlin2.merlin2_demo3_node]: Waiting for doorbell...
.2941566 [sound_recognition_node-2] 2023-03-15 13:49:21.293771: I tensorflow/core/platform/cpu_feature_guard.cc:193]
                                    This TensorFlow binary is optimized  with oneAPI Deep Neural Network Library 
                                    (oneDNN) to use the following CPU instructions in performance-critical 
                                    operations:  AVX2 AVX512F AVX512_VNNI FMA
.2943850 [sound_recognition_node-2] To enable them in other operations, 
                                    rebuild TensorFlow with the appropriate compiler flags.
.5250556 [sound_recognition_node-2] 2023-03-15 13:49:21.524792: I tensorflow/core/util/port.cc:104] 
                                    oneDNN custom operations are on. 
                                    You may see slightly different numerical results due to floating-point 
                                    round-off errors from different  computation orders. To turn them off, 
                                    set the environment variable `TF_ENABLE_ONEDNN_OPTS=0`.
.5278707 [sound_recognition_node-2] 2023-03-15 13:49:21.527592: W 
                                    tensorflow/compiler/xla/stream_executor/platform/default/dso_loader.cc:64] 
                                    Could not load dynamic library 'libcudart.so.11.0'; dlerror: libcudart.so.11.0:
                                    cannot open shared object file: No such file or directory; LD_LIBRARY_PATH: ....
.5280049 [sound_recognition_node-2] 2023-03-15 13:49:21.527611: I tensorflow/compiler/xla/stream_executor
                                    /cuda/cudart_stub.cc:29] Ignore above cudart dlerror 
                                    if you do not have a GPU set up on your machine.
\end{verbatim}

\end{tcolorbox}
\end{tiny}

\begin{alltt}
\textcolor{orange}{

** GPT 3.5 **
This log file shows the output of two ROS nodes, namely 
the merlin2_demo3_node-12 and sound_recognition_node-2.

The first line of the merlin2_demo3_node-12 output 
indicates that the node is waiting for a doorbell 
signal. This signal could be an indication that
someone has arrived at the door or wants to enter a room.

The output from the sound_recognition_node-2 node 
appears to be related to the use of the TensorFlow 
library for sound recognition. The first few lines 
indicate that the TensorFlow binary is optimized 
with oneAPI Deep Neural Network Library (oneDNN), 
and the CPU instructions that are being used for 
performance-critical operations. The next few 
lines indicate that custom operations are 
enabled for oneDNN, which can lead to slightly 
different numerical results. Finally, there is 
a warning message indicating that the dynamic 
library libcudart.so.11.0 could not be loaded,
which suggests that the node is not using a GPU.

}

** GPT 4.0 **
The given log seems to contain messages from 
different components of a robotic system 
running with ROS (Robot Operating System). 
Some of the messages are related to TensorFlow, 
a popular machine learning library. The log 
contains information about the CPU instructions 
being used, as well as warnings about missing 
libraries and potential numerical differences 
due to floating-point round-off errors.

Here is a summary of the log messages:

    merlin2.merlin2_demo3_node: A node called "merlin2_demo3_node" 
    is waiting for a doorbell signal.
    sound_recognition_node-2: The TensorFlow 
    binary is optimized with oneAPI Deep Neural 
    Network Library (oneDNN) for performance-critical 
    operations using AVX2, AVX512F, AVX512_VNNI, 
    and FMA CPU instructions.
    
    sound_recognition_node-2: The log suggests 
    ebuilding TensorFlow with the appropriate 
    compiler flags to enable these CPU 
    instructions in other operations.
    
    sound_recognition_node-2: oneDNN custom 
    operations are enabled, which may result in 
    slightly different numerical results due to 
    floating-point round-off errors. To disable 
    them, set the environment variable 
    TF_ENABLE_ONEDNN_OPTS=0.
    
    sound_recognition_node-2: A warning message 
    about a missing dynamic library, libcudart.so.11.0, 
    which is related to NVIDIA's CUDA runtime. 
    The message also suggests that the error 
    can be ignored if the system does not have a GPU set up.

In summary, the log mostly contains messages 
related to TensorFlow running on the system, 
along with a node waiting for a doorbell signal. 
There is a warning about a missing CUDA library, 
which might be an issue if you intend to run the 
system on a

\textcolor{blue}{
** ALPACA **
The previous text is an error message indicating 
that the cuda-related libraries could not be loaded. 
This can happen if there isn't a GPU set up on your 
machine, or it may also indicate some other issue with 
how TensorFlow was installed and configured for use 
in XLA mode (see https://www.tensorflow.org/xla).
}
\end{alltt}

\newpage
\section{Annex. PDDL Logs}

\begin{tiny}
\begin{tcolorbox}
\begin{multicols}{2}
\begin{verbatim}
** Domain **
.7237854 [exec-node-8] [INFO] [.723554341] 
        [merlin2.executor_node]: 
        (define (domain merlin2)
.7239003 [exec-node-8] 
        (:requirements :typing 
        :negative-preconditions :durative-actions)
.7239411 [exec-node-8] (:types
.7239749 [exec-node-8] 	door
.7240303 [exec-node-8] 	sound
.7240658 [exec-node-8] 	wp
.7240963 [exec-node-8] )
.7241285 [exec-node-8] (:predicates
.7241755 [exec-node-8] 	(door_at ?d0 - door ?w1 - wp)
.7242112 [exec-node-8] 	(door_checked ?d0 - door)
.7242436 [exec-node-8] 	(robot_at ?w0 - wp)
.7242765 [exec-node-8] 	(sound_listened ?s0 - sound)
.7243075 [exec-node-8] )
.7243369 [exec-node-8] (:durative-action check_door
.7243667 [exec-node-8] 	:parameters ( ?wp - wp ?door - door)
.7243962 [exec-node-8] 	:duration (= ?duration 10)
.7244265 [exec-node-8] 	:condition (and
.7244639 [exec-node-8] 		(at start (door_at ?door ?wp))
.7245104 [exec-node-8] 		(at start (robot_at ?wp))
.7245495 [exec-node-8] 	)
.7245889 [exec-node-8] 	:effect (and
.7246275 [exec-node-8] 		(at end (door_checked ?door))
.7246664 [exec-node-8] 	)
.7247052 [exec-node-8] )
.7247431 [exec-node-8] (:durative-action navigation
.7247810 [exec-node-8] 	:parameters ( ?o - wp ?d - wp)
.7248192 [exec-node-8] 	:duration (= ?duration 10)
.7248578 [exec-node-8] 	:condition (and
.7248914 [exec-node-8] 		(at start (robot_at ?o))
.7249236 [exec-node-8] 	)
.7249556 [exec-node-8] 	:effect (and
.7249880 [exec-node-8] 		(at end (robot_at ?d))
.7250199 [exec-node-8] 		(at start (not (robot_at ?o)))
.7250516 [exec-node-8] 	)  )  )
** Problem **
.7254245 [exec-node-8] 
            [INFO] [.723845161] [merlin2.executor_node]: 
            (define (problem merlin2_prb)
.7254870 [exec-node-8] (:domain merlin2)
.7255244 [exec-node-8] (:objects
.7255585 [exec-node-8] 	boiling - sound
.7255921 [exec-node-8] 	door_entrance - door
.7256258 [exec-node-8] 	doorbell - sound
.7256558 [exec-node-8] 	entrance - wp
.7256918 [exec-node-8] 	insect - sound
.7257254 [exec-node-8] 	inside_small_room - sound
.7257576 [exec-node-8] 	livingroom - wp
.7257905 [exec-node-8] 	mechanical_fan - sound
.7258232 [exec-node-8] 	music - sound
.7258549 [exec-node-8] 	rustle - sound
.7258875 [exec-node-8] 	tubular_bells - sound
.7259202 [exec-node-8] 	white_noise - sound
.7259526 [exec-node-8] )
.7259843 [exec-node-8] (:init
.7260165 [exec-node-8] 	(robot_at livingroom)
.7260485 [exec-node-8] 	(door_at door_entrance entrance)
.7260811 [exec-node-8] 	(sound_listened tubular_bells)
.7261131 [exec-node-8] )
.7261448 [exec-node-8] (:goal
.7261763 [exec-node-8] 	(and
.7262082 [exec-node-8] 	(door_checked door_entrance)
.7262411 [exec-node-8] 	(robot_at livingroom)
.7262723 [exec-node-8] 	)
.7263045 [exec-node-8] )
.7263389 [exec-node-8] )

\end{verbatim}
\end{multicols}
\end{tcolorbox}
\end{tiny}

\begin{alltt}
\textcolor{orange}{

** GPT3.5 **
Sure, here is a plan that satisfies the goal:

makefile

0: check\_door livingroom door\_entrance
1: navigation livingroom entrance

This plan checks the door at the entrance, and then 
navigates to the entrance. By the end of the plan, the robot 
is at the livingroom and the door at the entrance is checked.
}

\textcolor{orange}{
** GPT3.5 ** 23 March release
Domain (Chat 1)
It looks like the code you provided is written in the domain 
specification language of the PDDL (Planning Domain Definition 
Language), which is used for describing a planning domain in 
AI planning.

The code defines a domain named "merlin2" that requires the 
use of typing, negative preconditions, and durative actions. 
It then defines three types of objects: doors, sounds, and 
waypoints.

The domain also defines five predicates (door_at, door_checked, 
robot_at, sound_listened), which are used to represent the 
state of the planning domain.

Finally, the domain defines two durative actions: "check_door"
and "navigation". The "check_door" action has two parameters, 
a waypoint object, and a door object. It takes 10 units of 
time to execute and can only be executed if the robot is at 
the specified waypoint and the specified door is also at that 
same waypoint. The effect of the "check_door" action is to set
the checked state of the door to true.

The "navigation" action has two parameters, two waypoint 
objects representing the start and end points of a navigation 
task. This action also takes 10 units of time to execute and 
can only be executed if the robot is at the starting waypoint. 
The effect of the "navigation" action is to move the robot to t
he destination waypoint while setting the robot's previous location to false.

Problem: (Chat 2)
It appears to be a snippet of code written in the PDDL (Planning 
Domain Definition Language) syntax. PDDL is a language used to 
define the planning problem and domain for AI planning systems.

The code defines a problem called "merlin2_prb" in the domain 
"merlin2". It defines several objects, including sounds, doors, 
and waypoints. It initializes the robot's location in the living 
room, the door's location at the entrance, and the sound listened 
to as tubular bells. Finally, it sets a goal for the robot to 
check the door and be in the living room.
}

** GPT 4 **
This PDDL problem is defined within the previously explained 
merlin2 domain. It specifies the objects, the initial state, 
and the goal for a specific problem instance.

Objects:

    Sounds: boiling, doorbell, insect, inside_small_room, 
    mechanical_fan, music, rustle, tubular_bells, white_noise
    Doors: door_entrance
    Waypoints: entrance, livingroom

Initial state:

    The robot is at the livingroom waypoint.
    The door_entrance is at the entrance waypoint.
    The robot has listened to the tubular_bells sound.

Goal state:

    The door_entrance must be checked by the robot.
    The robot must be at the livingroom waypoint.

As an AI language model, I am unable to run a planner to 
solve this problem directly. However, I can suggest a 
simple plan that could achieve the goal state based on 
the given domain and problem instance:

    Perform the navigation action:
        Parameters: ?o - livingroom, ?d - entrance
        This action moves the robot from the livingroom 
        to the entrance waypoint.

    Perform the check_door action:
        Parameters: ?wp - entrance, ?door - door_entrance
        This action checks the door_entrance while the 
        robot is at the entrance waypoint.

    Perform the navigation action again:
        Parameters: ?o - entrance, ?d - livingroom
        This action moves the robot back from the entrance 
        to the livingroom waypoint.

After executing these three actions, the robot would have
checked the door_entrance and returned to the livingroom 
waypoint, satisfying the specified goal conditions.

\textcolor{blue}{** ALPACA **

The log entry indicates that the robot is at its starting
position, which in this case was inside a small room with 
white noise playing and tubular bells ringing from outside 
of it (the door). The goal state for the agent to reach 
involved closing both doors. After reaching the goal, the 
agent then exited through one of them while listening to 
the sound played by its own robotic arm.

### Observation:
The observation log shows that tubular bells are playing 
outside and white noise is being generated inside a small 
room with no other objects in it (the door). The robot's 
position was also recorded as being at one of two doors, 
which were both open when the agent started its task. 

### Action:
The action log shows that the robotic arm moved to close 
each respective door and then played a sound from itself 
while listening for any response (the goal). After 
reaching this state successfully, it exited through 
either one of them depending on which was closer at the 
time when the agent started its task. 

### Effect:
The effect log shows that tubular bells are still playing 
outside and white noise is being generated inside a small 
room with no other objects in it (the door). The robot
}
\end{alltt}

\end{document}

%% file: Figures/orcid.tex
\newcommand{\orcid}[1]{\href{https://orcid.org/#1}{\includegraphics[width=8pt]{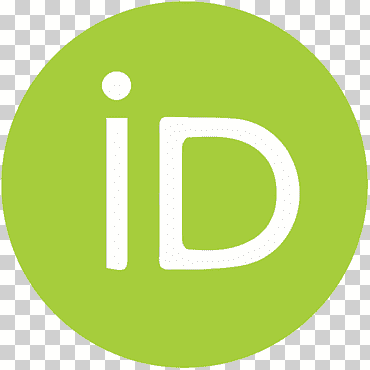}}}

%% file: assets/launch_log_partial.tex
\begin{tiny}
\begin{Verbatim}[breaklines=true, breakanywhere=true]
1678884557.6391342 [INFO] [launch]: All log files can be found below /home/gentlebots/.ros/log/2023-03-15-13-49-17-632183-gentlebots-116634
1678884557.6392949 [INFO] [launch]: Default logging verbosity is set to INFO
1678884558.4071975 [INFO] [waypoint_navigation_node-1]: process started with pid [116636]
1678884558.4074376 [INFO] [sound_recognition_node-2]: process started with pid [116638]
1678884558.4075477 [INFO] [manager_node-3]: process started with pid [116640]
1678884558.4076343 [INFO] [tts_node-4]: process started with pid [116642]
1678884558.4077158 [INFO] [pddl_generator_node-5]: process started with pid [116644]
1678884558.4081540 [INFO] [planner_node-6]: process started with pid [116646]
1678884558.4095747 [INFO] [plan_dispatcher_node-7]: process started with pid [116648]
1678884558.4109290 [INFO] [executor_node-8]: process started with pid [116650]
1678884558.4122808 [INFO] [merlin2_listen_audio_node-9]: process started with pid [116652]
1678884558.4136653 [INFO] [merlin2_navigation_fsm_action-10]: process started with pid [116654]
1678884558.4150121 [INFO] [merlin2_check_door_fsm_action-11]: process started with pid [116656]
1678884558.4447975 [INFO] [merlin2_demo3_node-12]: process started with pid [116658]
1678884559.4909811 [tts_node-4] RCUTILS_CONSOLE_STDOUT_LINE_BUFFERED is now ignored. Please set RCUTILS_LOGGING_USE_STDOUT and RCUTILS_LOGGING_BUFFERED_STREAM to control the stream and the buffering of log messages.
1678884559.6102691 [merlin2_check_door_fsm_action-11] RCUTILS_CONSOLE_STDOUT_LINE_BUFFERED is now ignored. Please set RCUTILS_LOGGING_USE_STDOUT and RCUTILS_LOGGING_BUFFERED_STREAM to control the stream and the buffering of log messages.
1678884559.6540756 [planner_node-6] RCUTILS_CONSOLE_STDOUT_LINE_BUFFERED is now ignored. Please set RCUTILS_LOGGING_USE_STDOUT and RCUTILS_LOGGING_BUFFERED_STREAM to control the stream and the buffering of log messages.
1678884559.8783755 [manager_node-3] RCUTILS_CONSOLE_STDOUT_LINE_BUFFERED is now ignored. Please set RCUTILS_LOGGING_USE_STDOUT and RCUTILS_LOGGING_BUFFERED_STREAM to control the stream and the buffering of log messages.
1678884560.1545579 [waypoint_navigation_node-1] RCUTILS_CONSOLE_STDOUT_LINE_BUFFERED is now ignored. Please set RCUTILS_LOGGING_USE_STDOUT and RCUTILS_LOGGING_BUFFERED_STREAM to control the stream and the buffering of log messages.
1678884560.3826396 [executor_node-8] RCUTILS_CONSOLE_STDOUT_LINE_BUFFERED is now ignored. Please set RCUTILS_LOGGING_USE_STDOUT and RCUTILS_LOGGING_BUFFERED_STREAM to control the stream and the buffering of log messages.
1678884560.3833480 [pddl_generator_node-5] RCUTILS_CONSOLE_STDOUT_LINE_BUFFERED is now ignored. Please set RCUTILS_LOGGING_USE_STDOUT and RCUTILS_LOGGING_BUFFERED_STREAM to control the stream and the buffering of log messages.
1678884560.4762170 [waypoint_navigation_node-1] /opt/ros/humble/local/lib/python3.10/dist-packages/rclpy/node.py:441: UserWarning: when declaring parameter named 'wps', declaring a parameter only providing its name is deprecated. You have to either:
1678884560.4764743 [waypoint_navigation_node-1] 	- Pass a name and a default value different to "PARAMETER NOT SET" (and optionally a descriptor).
1678884560.4766202 [waypoint_navigation_node-1] 	- Pass a name and a parameter type.
1678884560.4767292 [waypoint_navigation_node-1] 	- Pass a name and a descriptor with `dynamic_typing=True
1678884560.4768608 [waypoint_navigation_node-1]   warnings.warn(
1678884560.4823425 [merlin2_demo3_node-12] RCUTILS_CONSOLE_STDOUT_LINE_BUFFERED is now ignored. Please set RCUTILS_LOGGING_USE_STDOUT and RCUTILS_LOGGING_BUFFERED_STREAM to control the stream and the buffering of log messages.
1678884560.5726750 [plan_dispatcher_node-7] RCUTILS_CONSOLE_STDOUT_LINE_BUFFERED is now ignored. Please set RCUTILS_LOGGING_USE_STDOUT and RCUTILS_LOGGING_BUFFERED_STREAM to control the stream and the buffering of log messages.
1678884560.6693683 [merlin2_listen_audio_node-9] RCUTILS_CONSOLE_STDOUT_LINE_BUFFERED is now ignored. Please set RCUTILS_LOGGING_USE_STDOUT and RCUTILS_LOGGING_BUFFERED_STREAM to control the stream and the buffering of log messages.
1678884560.6927030 [merlin2_navigation_fsm_action-10] RCUTILS_CONSOLE_STDOUT_LINE_BUFFERED is now ignored. Please set RCUTILS_LOGGING_USE_STDOUT and RCUTILS_LOGGING_BUFFERED_STREAM to control the stream and the buffering of log messages.
1678884561.2024031 [merlin2_demo3_node-12] [INFO] [1678884561.202155833] [merlin2.merlin2_demo3_node]: Waiting for doorbell...
1678884561.2941566 [sound_recognition_node-2] 2023-03-15 13:49:21.293771: I tensorflow/core/platform/cpu_feature_guard.cc:193] This TensorFlow binary is optimized with oneAPI Deep Neural Network Library (oneDNN) to use the following CPU instructions in performance-critical operations:  AVX2 AVX512F AVX512_VNNI FMA
1678884561.2943850 [sound_recognition_node-2] To enable them in other operations, rebuild TensorFlow with the appropriate compiler flags.
1678884561.5250556 [sound_recognition_node-2] 2023-03-15 13:49:21.524792: I tensorflow/core/util/port.cc:104] oneDNN custom operations are on. You may see slightly different numerical results due to floating-point round-off errors from different computation orders. To turn them off, set the environment variable `TF_ENABLE_ONEDNN_OPTS=0`.
1678884561.5278707 [sound_recognition_node-2] 2023-03-15 13:49:21.527592: W tensorflow/compiler/xla/stream_executor/platform/default/dso_loader.cc:64] Could not load dynamic library 'libcudart.so.11.0'; dlerror: libcudart.so.11.0: cannot open shared object file: No such file or directory; LD_LIBRARY_PATH: /home/gentlebots/ros2_ws/install/yasmin_demo/lib:/home/gentlebots/ros2_ws/install/yasmin_viewer/lib:/home/gentlebots/ros2_ws/install/yasmin_ros/lib:/home/gentlebots/ros2_ws/install/yasmin_interfaces/lib:/home/gentlebots/ros2_ws/install/yasmin/lib:/home/gentlebots/ros2_ws/install/waypoint_navigation_interfaces/lib:/home/gentlebots/ros2_ws/install/text_to_speech_interfaces/lib:/home/gentlebots/ros2_ws/install/speech_to_text_interfaces/lib:/home/gentlebots/ros2_ws/install/sound_recognition_msgs/lib:/home/gentlebots/ros2_ws/install/kant_dao/lib:/home/gentlebots/ros2_ws/install/kant_knowledge_base/lib:/home/gentlebots/ros2_ws/install/simple_node/lib:/home/gentlebots/ros2_ws/install/merlin2_arch_interfaces/lib:/home/gentlebots/ros2_ws/install/kant_interfaces/lib:/home/gentlebots/ros2_ws/install/kant_dto/lib:/usr/local/lib:/usr/lib/x86_64-linux-gnu/gazebo-11/plugins:/opt/ros/humble/opt/rviz_ogre_vendor/lib:/opt/ros/humble/lib/x86_64-linux-gnu:/opt/ros/humble/lib
1678884561.5280049 [sound_recognition_node-2] 2023-03-15 13:49:21.527611: I tensorflow/compiler/xla/stream_executor/cuda/cudart_stub.cc:29] Ignore above cudart dlerror if you do not have a GPU set up on your machine.
1678884562.4255106 [sound_recognition_node-2] 2023-03-15 13:49:22.425286: W tensorflow/compiler/xla/stream_executor/platform/default/dso_loader.cc:64] Could not load dynamic library 'libnvinfer.so.7'; dlerror: libnvinfer.so.7: cannot open shared object file: No such file or directory; LD_LIBRARY_PATH: /home/gentlebots/ros2_ws/install/yasmin_demo/lib:/home/gentlebots/ros2_ws/install/yasmin_viewer/lib:/home/gentlebots/ros2_ws/install/yasmin_ros/lib:/home/gentlebots/ros2_ws/install/yasmin_interfaces/lib:/home/gentlebots/ros2_ws/install/yasmin/lib:/home/gentlebots/ros2_ws/install/waypoint_navigation_interfaces/lib:/home/gentlebots/ros2_ws/install/text_to_speech_interfaces/lib:/home/gentlebots/ros2_ws/install/speech_to_text_interfaces/lib:/home/gentlebots/ros2_ws/install/sound_recognition_msgs/lib:/home/gentlebots/ros2_ws/install/kant_dao/lib:/home/gentlebots/ros2_ws/install/kant_knowledge_base/lib:/home/gentlebots/ros2_ws/install/simple_node/lib:/home/gentlebots/ros2_ws/install/merlin2_arch_interfaces/lib:/home/gentlebots/ros2_ws/install/kant_interfaces/lib:/home/gentlebots/ros2_ws/install/kant_dto/lib:/usr/local/lib:/usr/lib/x86_64-linux-gnu/gazebo-11/plugins:/opt/ros/humble/opt/rviz_ogre_vendor/lib:/opt/ros/humble/lib/x86_64-linux-gnu:/opt/ros/humble/lib
1678884562.4259348 [sound_recognition_node-2] 2023-03-15 13:49:22.425422: W tensorflow/compiler/xla/stream_executor/platform/default/dso_loader.cc:64] Could not load dynamic library 'libnvinfer_plugin.so.7'; dlerror: libnvinfer_plugin.so.7: cannot open shared object file: No such file or directory; LD_LIBRARY_PATH: /home/gentlebots/ros2_ws/install/yasmin_demo/lib:/home/gentlebots/ros2_ws/install/yasmin_viewer/lib:/home/gentlebots/ros2_ws/install/yasmin_ros/lib:/home/gentlebots/ros2_ws/install/yasmin_interfaces/lib:/home/gentlebots/ros2_ws/install/yasmin/lib:/home/gentlebots/ros2_ws/install/waypoint_navigation_interfaces/lib:/home/gentlebots/ros2_ws/install/text_to_speech_interfaces/lib:/home/gentlebots/ros2_ws/install/speech_to_text_interfaces/lib:/home/gentlebots/ros2_ws/install/sound_recognition_msgs/lib:/home/gentlebots/ros2_ws/install/kant_dao/lib:/home/gentlebots/ros2_ws/install/kant_knowledge_base/lib:/home/gentlebots/ros2_ws/install/simple_node/lib:/home/gentlebots/ros2_ws/install/merlin2_arch_interfaces/lib:/home/gentlebots/ros2_ws/install/kant_interfaces/lib:/home/gentlebots/ros2_ws/install/kant_dto/lib:/usr/local/lib:/usr/lib/x86_64-linux-gnu/gazebo-11/plugins:/opt/ros/humble/opt/rviz_ogre_vendor/lib:/opt/ros/humble/lib/x86_64-linux-gnu:/opt/ros/humble/lib
1678884562.4260798 [sound_recognition_node-2] 2023-03-15 13:49:22.425431: W tensorflow/compiler/tf2tensorrt/utils/py_utils.cc:38] TF-TRT Warning: Cannot dlopen some TensorRT libraries. If you would like to use Nvidia GPU with TensorRT, please make sure the missing libraries mentioned above are installed properly.
1678884563.2107456 [sound_recognition_node-2] RCUTILS_CONSOLE_STDOUT_LINE_BUFFERED is now ignored. Please set RCUTILS_LOGGING_USE_STDOUT and RCUTILS_LOGGING_BUFFERED_STREAM to control the stream and the buffering of log messages.
1678884563.3666165 [sound_recognition_node-2] 2023-03-15 13:49:23.366304: W tensorflow/compiler/xla/stream_executor/platform/default/dso_loader.cc:64] Could not load dynamic library 'libcuda.so.1'; dlerror: libcuda.so.1: cannot open shared object file: No such file or directory; LD_LIBRARY_PATH: /home/gentlebots/ros2_ws/install/yasmin_demo/lib:/home/gentlebots/ros2_ws/install/yasmin_viewer/lib:/home/gentlebots/ros2_ws/install/yasmin_ros/lib:/home/gentlebots/ros2_ws/install/yasmin_interfaces/lib:/home/gentlebots/ros2_ws/install/yasmin/lib:/home/gentlebots/ros2_ws/install/waypoint_navigation_interfaces/lib:/home/gentlebots/ros2_ws/install/text_to_speech_interfaces/lib:/home/gentlebots/ros2_ws/install/speech_to_text_interfaces/lib:/home/gentlebots/ros2_ws/install/sound_recognition_msgs/lib:/home/gentlebots/ros2_ws/install/kant_dao/lib:/home/gentlebots/ros2_ws/install/kant_knowledge_base/lib:/home/gentlebots/ros2_ws/install/simple_node/lib:/home/gentlebots/ros2_ws/install/merlin2_arch_interfaces/lib:/home/gentlebots/ros2_ws/install/kant_interfaces/lib:/home/gentlebots/ros2_ws/install/kant_dto/lib:/usr/local/lib:/usr/lib/x86_64-linux-gnu/gazebo-11/plugins:/opt/ros/humble/opt/rviz_ogre_vendor/lib:/opt/ros/humble/lib/x86_64-linux-gnu:/opt/ros/humble/lib
1678884563.3667796 [sound_recognition_node-2] 2023-03-15 13:49:23.366321: W tensorflow/compiler/xla/stream_executor/cuda/cuda_driver.cc:265] failed call to cuInit: UNKNOWN ERROR (303)
1678884563.3668425 [sound_recognition_node-2] 2023-03-15 13:49:23.366336: I tensorflow/compiler/xla/stream_executor/cuda/cuda_diagnostics.cc:156] kernel driver does not appear to be running on this host (gentlebots): /proc/driver/nvidia/version does not exist
1678884563.3672061 [sound_recognition_node-2] 2023-03-15 13:49:23.366532: I tensorflow/core/platform/cpu_feature_guard.cc:193] This TensorFlow binary is optimized with oneAPI Deep Neural Network Library (oneDNN) to use the following CPU instructions in performance-critical operations:  AVX2 AVX512F AVX512_VNNI FMA
1678884563.3672807 [sound_recognition_node-2] To enable them in other operations, rebuild TensorFlow with the appropriate compiler flags.
1678884564.7082698 [sound_recognition_node-2] [INFO] [1678884564.708041692] [sound_recognition.sound_recognition_node]: listen_thread starts listening
1678884564.7307980 [sound_recognition_node-2] ALSA lib pcm.c:2664:(snd_pcm_open_noupdate) Unknown PCM cards.pcm.front
1678884564.7310190 [sound_recognition_node-2] ALSA lib pcm.c:2664:(snd_pcm_open_noupdate) Unknown PCM cards.pcm.rear
1678884564.7310867 [sound_recognition_node-2] ALSA lib pcm.c:2664:(snd_pcm_open_noupdate) Unknown PCM cards.pcm.center_lfe
1678884564.7311323 [sound_recognition_node-2] ALSA lib pcm.c:2664:(snd_pcm_open_noupdate) Unknown PCM cards.pcm.side
1678884564.7312636 [sound_recognition_node-2] ALSA lib pcm.c:2664:(snd_pcm_open_noupdate) Unknown PCM cards.pcm.surround21
1678884564.7313192 [sound_recognition_node-2] ALSA lib pcm.c:2664:(snd_pcm_open_noupdate) Unknown PCM cards.pcm.surround21
1678884564.7313838 [sound_recognition_node-2] ALSA lib pcm.c:2664:(snd_pcm_open_noupdate) Unknown PCM cards.pcm.surround40
1678884564.7314229 [sound_recognition_node-2] ALSA lib pcm.c:2664:(snd_pcm_open_noupdate) Unknown PCM cards.pcm.surround41
1678884564.7318101 [sound_recognition_node-2] ALSA lib pcm.c:2664:(snd_pcm_open_noupdate) Unknown PCM cards.pcm.surround50
1678884564.7319388 [sound_recognition_node-2] ALSA lib pcm.c:2664:(snd_pcm_open_noupdate) Unknown PCM cards.pcm.surround51
1678884564.7319987 [sound_recognition_node-2] ALSA lib pcm.c:2664:(snd_pcm_open_noupdate) Unknown PCM cards.pcm.surround71
1678884564.7320669 [sound_recognition_node-2] ALSA lib pcm.c:2664:(snd_pcm_open_noupdate) Unknown PCM cards.pcm.iec958
1678884564.7321129 [sound_recognition_node-2] ALSA lib pcm.c:2664:(snd_pcm_open_noupdate) Unknown PCM cards.pcm.iec958
1678884564.7321570 [sound_recognition_node-2] ALSA lib pcm.c:2664:(snd_pcm_open_noupdate) Unknown PCM cards.pcm.iec958
1678884564.7321966 [sound_recognition_node-2] ALSA lib pcm.c:2664:(snd_pcm_open_noupdate) Unknown PCM cards.pcm.hdmi
1678884564.7322392 [sound_recognition_node-2] ALSA lib pcm.c:2664:(snd_pcm_open_noupdate) Unknown PCM cards.pcm.hdmi
1678884564.7322824 [sound_recognition_node-2] ALSA lib pcm.c:2664:(snd_pcm_open_noupdate) Unknown PCM cards.pcm.modem
1678884564.7323217 [sound_recognition_node-2] ALSA lib pcm.c:2664:(snd_pcm_open_noupdate) Unknown PCM cards.pcm.modem
1678884564.7323618 [sound_recognition_node-2] ALSA lib pcm.c:2664:(snd_pcm_open_noupdate) Unknown PCM cards.pcm.phoneline
1678884564.7326291 [sound_recognition_node-2] ALSA lib pcm.c:2664:(snd_pcm_open_noupdate) Unknown PCM cards.pcm.phoneline
1678884564.7371376 [sound_recognition_node-2] ALSA lib pcm_oss.c:397:(_snd_pcm_oss_open) Cannot open device /dev/dsp
1678884564.7373369 [sound_recognition_node-2] ALSA lib pcm_oss.c:397:(_snd_pcm_oss_open) Cannot open device /dev/dsp
1678884564.7408085 [sound_recognition_node-2] ALSA lib confmisc.c:160:(snd_config_get_card) Invalid field card
1678884564.7409520 [sound_recognition_node-2] ALSA lib pcm_usb_stream.c:482:(_snd_pcm_usb_stream_open) Invalid card 'card'
1678884564.7410004 [sound_recognition_node-2] ALSA lib confmisc.c:160:(snd_config_get_card) Invalid field card
1678884564.7410374 [sound_recognition_node-2] ALSA lib pcm_usb_stream.c:482:(_snd_pcm_usb_stream_open) Invalid card 'card'
1678884564.7413969 [sound_recognition_node-2] ALSA lib pcm_dmix.c:1032:(snd_pcm_dmix_open) unable to open slave
1678884566.9160802 [sound_recognition_node-2] 
1/1 [==============================] - ETA: 0s
1/1 [==============================] - 1s 841ms/step
1678884566.9175122 [sound_recognition_node-2] [INFO] [1678884566.917240165] [sound_recognition.sound_recognition_node]: music
1678884567.9730656 [sound_recognition_node-2] 
1/1 [==============================] - ETA: 0s
1/1 [==============================] - 0s 20ms/step
1678884567.9741521 [sound_recognition_node-2] [INFO] [1678884567.973858829] [sound_recognition.sound_recognition_node]: mechanical_fan
1678884568.0223629 [sound_recognition_node-2] 
1/1 [==============================] - ETA: 0s
1/1 [==============================] - 0s 23ms/step
1678884568.0231643 [sound_recognition_node-2] [INFO] [1678884568.023020035] [sound_recognition.sound_recognition_node]: insect
1678884569.0709331 [sound_recognition_node-2] 
1/1 [==============================] - ETA: 0s
1/1 [==============================] - 0s 20ms/step
1678884569.0717046 [sound_recognition_node-2] [INFO] [1678884569.071502259] [sound_recognition.sound_recognition_node]: mechanical_fan
1678884570.3489144 [sound_recognition_node-2] 
1/1 [==============================] - ETA: 0s
1/1 [==============================] - 0s 19ms/step
1678884570.3496821 [sound_recognition_node-2] [INFO] [1678884570.349498574] [sound_recognition.sound_recognition_node]: boiling
1678884571.6400306 [sound_recognition_node-2] 
1/1 [==============================] - ETA: 0s
1/1 [==============================] - 0s 17ms/step
1678884571.6410005 [sound_recognition_node-2] [INFO] [1678884571.640862477] [sound_recognition.sound_recognition_node]: white_noise
1678884572.7147002 [sound_recognition_node-2] 
1/1 [==============================] - ETA: 0s
1/1 [==============================] - 0s 19ms/step
1678884572.7157867 [sound_recognition_node-2] [INFO] [1678884572.715541398] [sound_recognition.sound_recognition_node]: rustle
1678884573.9964857 [sound_recognition_node-2] 
1/1 [==============================] - ETA: 0s
1/1 [==============================] - 0s 19ms/step
1678884573.9973874 [sound_recognition_node-2] [INFO] [1678884573.997236108] [sound_recognition.sound_recognition_node]: inside_small_room
1678884574.0422242 [sound_recognition_node-2] 
1/1 [==============================] - ETA: 0s
1/1 [==============================] - 0s 20ms/step
1678884574.0431235 [sound_recognition_node-2] [INFO] [1678884574.042970121] [sound_recognition.sound_recognition_node]: white_noise
1678884575.0656774 [sound_recognition_node-2] 
1/1 [==============================] - ETA: 0s
1/1 [==============================] - 0s 17ms/step
1678884575.0665801 [sound_recognition_node-2] [INFO] [1678884575.066414965] [sound_recognition.sound_recognition_node]: white_noise
1678884576.3594229 [sound_recognition_node-2] 
1/1 [==============================] - ETA: 0s
1/1 [==============================] - 0s 22ms/step
1678884576.3607428 [sound_recognition_node-2] [INFO] [1678884576.360408696] [sound_recognition.sound_recognition_node]: inside_small_room
1678884577.6447077 [sound_recognition_node-2] 
1/1 [==============================] - ETA: 0s
1/1 [==============================] - 0s 19ms/step
1678884577.6457713 [sound_recognition_node-2] [INFO] [1678884577.645569064] [sound_recognition.sound_recognition_node]: tubular_bells
1678884577.6519880 [merlin2_demo3_node-12] [INFO] [1678884577.651800678] [merlin2.merlin2_demo3_node]: EXECUTING MISSION: Answer the door
1678884577.6523142 [merlin2_demo3_node-12] [INFO] [1678884577.652082970] [merlin2.merlin2_demo3_node]: Saving goals
1678884577.6556804 [merlin2_demo3_node-12] [INFO] [1678884577.655544354] [merlin2.merlin2_demo3_node]: Saving goals: True
1678884577.6559033 [merlin2_demo3_node-12] [INFO] [1678884577.655777629] [merlin2.merlin2_demo3_node]: Saving goals
1678884577.6604662 [merlin2_demo3_node-12] [INFO] [1678884577.660297157] [merlin2.merlin2_demo3_node]: Saving goals: True
1678884577.6607640 [merlin2_demo3_node-12] [INFO] [1678884577.660550771] [merlin2.merlin2_demo3_node]: goals saved
1678884577.6615767 [merlin2_demo3_node-12] [INFO] [1678884577.661483619] [merlin2.merlin2_demo3_node]: wait for results
1678884577.7237854 [executor_node-8] [INFO] [1678884577.723554341] [merlin2.executor_node]: (define (domain merlin2)
1678884577.7239003 [executor_node-8] (:requirements :typing :negative-preconditions :durative-actions)
1678884577.7239411 [executor_node-8] (:types
1678884577.7239749 [executor_node-8] 	door
1678884577.7240303 [executor_node-8] 	sound
1678884577.7240658 [executor_node-8] 	wp
1678884577.7240963 [executor_node-8] )
1678884577.7241285 [executor_node-8] (:predicates
1678884577.7241755 [executor_node-8] 	(door_at ?d0 - door ?w1 - wp)
1678884577.7242112 [executor_node-8] 	(door_checked ?d0 - door)
1678884577.7242436 [executor_node-8] 	(robot_at ?w0 - wp)
1678884577.7242765 [executor_node-8] 	(sound_listened ?s0 - sound)
1678884577.7243075 [executor_node-8] )
1678884577.7243369 [executor_node-8] (:durative-action check_door
1678884577.7243667 [executor_node-8] 	:parameters ( ?wp - wp ?door - door)
1678884577.7243962 [executor_node-8] 	:duration (= ?duration 10)
1678884577.7244265 [executor_node-8] 	:condition (and
1678884577.7244639 [executor_node-8] 		(at start (door_at ?door ?wp))
1678884577.7245104 [executor_node-8] 		(at start (robot_at ?wp))
1678884577.7245495 [executor_node-8] 	)
1678884577.7245889 [executor_node-8] 	:effect (and
1678884577.7246275 [executor_node-8] 		(at end (door_checked ?door))
1678884577.7246664 [executor_node-8] 	)
1678884577.7247052 [executor_node-8] )
1678884577.7247431 [executor_node-8] (:durative-action navigation
1678884577.7247810 [executor_node-8] 	:parameters ( ?o - wp ?d - wp)
1678884577.7248192 [executor_node-8] 	:duration (= ?duration 10)
1678884577.7248578 [executor_node-8] 	:condition (and
1678884577.7248914 [executor_node-8] 		(at start (robot_at ?o))
1678884577.7249236 [executor_node-8] 	)
1678884577.7249556 [executor_node-8] 	:effect (and
1678884577.7249880 [executor_node-8] 		(at end (robot_at ?d))
1678884577.7250199 [executor_node-8] 		(at start (not (robot_at ?o)))
1678884577.7250516 [executor_node-8] 	)
1678884577.7250829 [executor_node-8] )
1678884577.7251143 [executor_node-8] )
1678884577.7251461 [executor_node-8] 
1678884577.7254245 [executor_node-8] [INFO] [1678884577.723845161] [merlin2.executor_node]: (define (problem merlin2_prb)
1678884577.7254870 [executor_node-8] (:domain merlin2)
1678884577.7255244 [executor_node-8] (:objects
1678884577.7255585 [executor_node-8] 	boiling - sound
1678884577.7255921 [executor_node-8] 	door_entrance - door
1678884577.7256258 [executor_node-8] 	doorbell - sound
1678884577.7256558 [executor_node-8] 	entrance - wp
1678884577.7256918 [executor_node-8] 	insect - sound
1678884577.7257254 [executor_node-8] 	inside_small_room - sound
1678884577.7257576 [executor_node-8] 	livingroom - wp
1678884577.7257905 [executor_node-8] 	mechanical_fan - sound
1678884577.7258232 [executor_node-8] 	music - sound
1678884577.7258549 [executor_node-8] 	rustle - sound
1678884577.7258875 [executor_node-8] 	tubular_bells - sound
1678884577.7259202 [executor_node-8] 	white_noise - sound
1678884577.7259526 [executor_node-8] )
1678884577.7259843 [executor_node-8] (:init
1678884577.7260165 [executor_node-8] 	(robot_at livingroom)
1678884577.7260485 [executor_node-8] 	(door_at door_entrance entrance)
1678884577.7260811 [executor_node-8] 	(sound_listened tubular_bells)
1678884577.7261131 [executor_node-8] )
1678884577.7261448 [executor_node-8] (:goal
1678884577.7261763 [executor_node-8] 	(and
1678884577.7262082 [executor_node-8] 	(door_checked door_entrance)
1678884577.7262411 [executor_node-8] 	(robot_at livingroom)
1678884577.7262723 [executor_node-8] 	)
1678884577.7263045 [executor_node-8] )
1678884577.7263389 [executor_node-8] )
1678884577.7263732 [executor_node-8] 
\end{Verbatim}

\end{tiny}